%% file: bmvc_review.tex
\documentclass{bmvc2k}
\usepackage{amssymb}
\usepackage{xurl}

\makeatletter
\g@addto@macro{\UrlBreaks}{\do\-}
\makeatother

\title{Simplified Cross-Modal Calibration for Heterogeneous Event-RGB Stereo Systems}

\addauthor{Nico Hessenthaler}{nico.hessenthaler@hs-heilbronn.de}{1}
\addauthor{Adam T. Müller}{adam-theo.mueller@hs-heilbronn.de}{1}
\addauthor{Nicolaj C. Stache}{nicolaj.stache@hs-heilbronn.de}{1}

\addinstitution{
 Center for Machine Learning\\
 Heilbronn University of Applied Sciences\\
 Heilbronn, Germany
}

\runninghead{Hessenthaler, Müller, Stache}{Simplified Event-RGB Calibration}

\def\etal{\emph{et al}\bmvaOneDot}

\hypersetup{
    pdftitle={Simplified Cross-Modal Calibration for Heterogeneous Event-RGB Stereo Systems},
    pdfauthor={Nico Hessenthaler, Adam T. Müller, Nicolaj C. Stache}
}

\begin{document}

\maketitle

\input{sec/0_abstract}    
\input{sec/1_intro}

\input{sec/2_relWork}
\input{sec/3_Method}
\input{sec/4_experiments}
\input{sec/5_limitations}

\input{sec/6_conclusion}

\newpage
\section*{Acknowledgements}
In part, this research was funded by the AI-TRAQC Project – Regional Innovation Center Artificial Intelligence Training \& Qualification Campus (RegioWIN 2472019), supported by the European Regional Development Fund (Europäischer Fonds für Regionale Entwicklung - EFRE).

\bibliography{Cross_Modal_Stereo_Calibration}

\newpage
\appendix
\input{sec/appendix}

\end{document}

%% file: sec/0_abstract.tex
\begin{abstract}
Accurate extrinsic calibration between event-based and frame-based cameras remains a practical bottleneck for heterogeneous stereo systems. Existing approaches often require sensor or target motion, precise synchronization, or computationally expensive event-to-image reconstruction. We propose a simple, motion-free cross-modal calibration framework that uses a temporally modulated, blended ChArUco target presented on standard consumer displays. By alternating between the original pattern and a partially blended version, the target reliably triggers events while remaining continuously observable to a frame-based camera, avoiding blank frames and reducing synchronization constraints to a coarse, trigger-based alignment. We discretize events into frames coarsely aligned with the RGB images, apply lightweight denoising, and perform ChArUco-based intrinsic and stereo extrinsic calibration. Extensive experiments assess robustness to blending opacity, display brightness, external illumination, viewing angle, and handheld acquisition. Compared to the strongest motion-based reference (E2Calib + Kalibr) and a non-motion-based reference (Plasberg et al.), our approach reduces the mean reprojection error by $44\%$ and $6\%$, respectively, while substantially simplifying the calibration procedure. Finally, we demonstrate practical utility in a robotic eye-to-hand calibration case study, showing consistent transformations and stable downstream geometric measurements even under partial occlusions. Code is publicly available at \url{https://github.com/nhessenthaler/simple-evrgb-cal}.
\end{abstract}

%% file: sec/1_intro.tex
\section{Introduction}
\label{sec:intro}

%



Accurate cross-modal calibration remains a bottleneck for the effective integration of event-based and frame-based cameras in heterogeneous sensor systems~\cite{cocheteux_muli-ev_2024,song_calibration_2018}. Similar to monocular event camera calibration~\cite{chen_ekalibr_2025}, cross-modal approaches can generally be categorized into motion-based and non-motion-based (static) methodologies.

Motion-based techniques~\cite{muglikar_how_2021,salah_e-calib_2024} trigger events through physical movement, but frequently demand computationally expensive neural image reconstructions or intricate algorithms to mitigate inherent motion blur. Conversely, static methods~\cite{gossard_ewand_2024,plasberg_intrinsic_2022} generate events by modulating light intensity, thus avoiding motion blur. However, these approaches typically rely on specialized hardware and require strict temporal synchronization to prevent the frame-based camera from losing tracking features during blank display cycles~\cite{bertogalli_one_2025}. Overcoming these practical complexities is highly motivated by dynamic application scenarios~\cite{vinod2025sebvs}, such as robotic hand-eye calibration~\cite{tsai_new_1989}. These systems require robust and continuous geometric references even when the target is partially occluded by a robotic manipulator.

To address these limitations, we introduce a straightforward, motion-free calibration framework that utilizes a shared, temporally modulated visual stimulus. By continuously alternating a ChArUco~\cite{garrido-jurado_automatic_2014} calibration pattern with a partially blended state on a standard consumer monitor, our method reliably induces the log-intensity changes necessary for event generation. Crucially, the target remains continuously visible to the RGB sensor throughout the cycle. This overlay methodology requires only a coarse, trigger-based alignment rather than strict hardware-level synchronization, while eliminating blank frame management, or computationally expensive event-to-image reconstructions. This substantially simplifies the calibration procedure for real-world deployment.

Our main contributions are as follows: \textbf{(1)} A novel cross-modal calibration methodology based on temporal image blending that eliminates the need for specialized hardware and complex image reconstruction, while reducing the synchronization requirement to a coarse alignment. \textbf{(2)} Comprehensive experimental validation demonstrating high statistical stability and robustness against varying display brightness, external illumination disturbances, viewing angle, and handheld acquisition, while reducing the mean reprojection error by $44\%$ and $6\%$ compared to the strongest evaluated motion-based and non-motion-based approaches, respectively. \textbf{(3)} A practical demonstration of the framework's utility in a robotic eye-to-hand calibration case study, achieving stable geometric measurements and consistent spatial transformations even in the presence of partial occlusions.


%% file: sec/2_relWork.tex
\section{Related Work}
\label{sec:relWork}


Building upon the core distinction between motion-based and static calibration methodologies, both paradigms fundamentally aim to trigger a continuous event stream by inducing pixel-level intensity changes. Beyond these two dominant categories, specialized frameworks such as EPCNet~\cite{sealy_phelan_epcnet_2025} and TUMTraf Event~\cite{cres_tumtraf_2024, cres_targetless_2023} address cross-modal calibration in application-specific contexts, particularly for intelligent transportation systems. An additional line of work aims to simplify cross-modal calibration by leveraging auxiliary sensing modalities. For instance, Dubeau \etal~\cite{dubeau_rgb-d-e_2020} utilize the Active Pixel Sensor (APS) frames of hybrid event cameras to enable calibration via conventional frame-based techniques.

\subsection{Motion-Based Calibration Approaches}
The dominant paradigm in the field applies relative motion between the sensor and a calibration target to trigger the event stream~\cite{huang_dynamic_2021, salah_e-calib_2024, chen_ekalibr_2025}. A prominent work in this category is the E2Calib framework proposed by Muglikar~\etal~\cite{muglikar_how_2021}, which utilizes neural-network-based image reconstruction (E2VID~\cite{rebecq_high_2021}) to convert asynchronous event streams into dense intensity frames. By reconstructing these intensity frames, standard computer vision toolboxes can be employed for both intrinsic and extrinsic calibration parameter estimation using a moving standard checkerboard pattern. However, Hu~\etal~\cite{hu_dynamic_2024} argue that reconstruction-based pipelines suffer from limited edge sharpness and reconstruction noise. To address these deficiencies, they propose a reconstruction-free approach. By utilizing a point-grid target, their method performs direct circle detection within the event stream, later matching the data with the frame-based camera using a sliding window time matching. Wang~\etal~\cite{wang_ef-calib_2024} employ a similar methodology but achieve superior subpixel accuracy through an optimized calibration target and an improved feature extraction pipeline. Similarly, Zhang~\etal~\cite{zhang_spatiotemporal_2025} refine motion-based cross-modal calibration by introducing motion-attracted event alignment, which leverages local consistency to mitigate motion blur in accumulated event frames.

Ultimately, these methodologies often present practical challenges for real-world deployment. They typically require complex neural reconstruction, precise synchronization, or sophisticated
motion-compensation algorithms to counteract motion blur. In contrast, our motion-free approach sidesteps these requirements, eliminating the dependency on dynamic feature triggering and the associated spatiotemporal alignment complexities.

\subsection{Non-Motion-Based Calibration Approaches}
Static calibration methodologies generate events by modulating light intensity rather than through physical motion, typically utilizing blinking LEDs ~\cite{gossard_ewand_2024, dominguez-morales_bio-inspired_2019, zi_flexible_2024, chen_novel_2020} or displays~\cite{cai_accurate_2024, wang_motion-error-free_2024}.

Gossard~\etal~\cite{gossard_ewand_2024} employ an asymmetric marker with frequency-modulated LEDs to distinguish features in the event stream while maintaining visibility for RGB cameras in low-light settings. However, this approach requires specialized hardware and lacks robustness in standard ambient lighting. Moreover, Bertogalli~\etal~\cite{bertogalli_one_2025} utilize a custom calibration cube integrating blinking LEDs, ChArUco~\cite{garrido-jurado_automatic_2014} patterns, and planar surfaces for RGB-event-LiDAR alignment. While effective, the reliance on precisely manufactured, specialized hardware limits its accessibility for general applications.

To leverage consumer hardware, Plasberg~\etal~\cite{plasberg_intrinsic_2022} utilize a binary blinking monitor behind a metallic dot-grid for thermal-event-RGB calibration. However, the authors omit any discussion of temporal synchronization between display and frame-based camera. A binary blinking mechanism may create a critical contrast failure. When the display cycles to black, the absence of contrast between the dark screen and the metallic grid can render the features undetectable to the frame-based camera. 

Our overlay-based methodology addresses these limitations, enabling high-contrast feature detection across modalities while reducing the operational complexity of cross-modal calibration. It eliminates the need for specialized hardware and intricate target designs, reduces synchronization constraints and preserves the advantages of static methods.

%% file: sec/3_Method.tex
\section{Methodology}
\label{sec:Method}

\begin{figure}[t]
\centering
\includegraphics[width=1.0\linewidth]{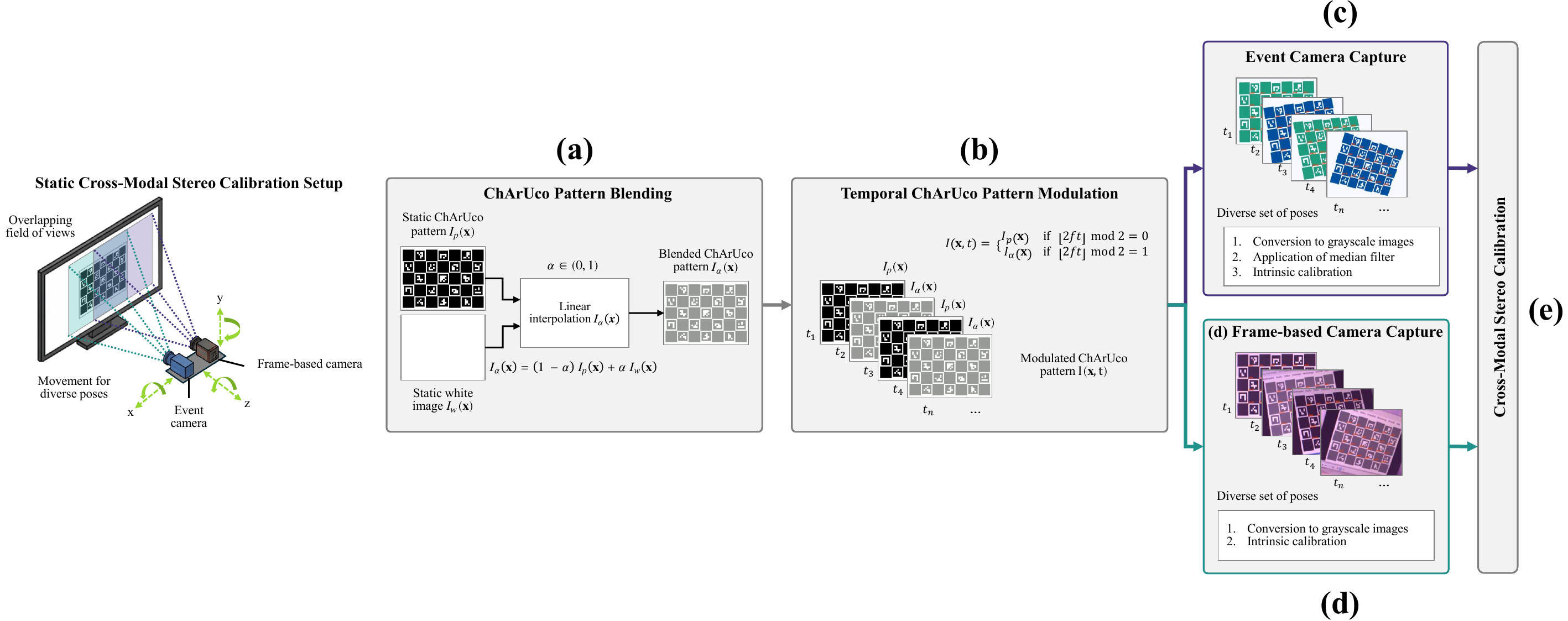}
\caption{Pipeline for cross‑modal stereo calibration using a temporally modulated ChArUco pattern. A pattern of blended images is generated via linear interpolation and modulated over time, enabling robust detection in both frame‑based and event cameras.}
\label{fig:method_overview}
\end{figure}
Our proposed method addresses cross-modal calibration between event-based and frame-based cameras by introducing a shared, temporally controlled visual stimulus that is observable in both sensing modalities. The core idea is to actively modulate a calibration pattern such that it simultaneously induces sufficient intensity changes for the event camera while remaining reliably detectable in the frame-based domain. This enables consistent feature extraction without relying on relative motion or precise temporal synchronization. The resulting signals are preprocessed independently to extract calibration features and estimate the intrinsic parameters of each sensor, before being jointly utilized for subsequent stereo calibration. A schematic overview of the full pipeline is shown in Fig.~\ref{fig:method_overview}, with detailed descriptions of each component provided in the following sections.


\paragraph{Calibration Target.}
In contrast to the standard checkerboard or point-grid patterns frequently employed in event-based literature, the proposed method utilizes a ChArUco-based calibration target, as illustrated in Fig.~\ref{fig:method_overview}~(a). While ChArUco patterns are already firmly established as a standard for frame-based camera calibration, their application to event-based sensors remains emerging. However, prior research has proven their efficacy for robust detection~\cite{zhang_improved_2023} and pose estimation~\cite{mai_pose_2023} in event-based vision, making them an ideal candidate for bridging the modality gap in calibrating heterogeneous stereo systems. The unique identification of marker IDs provides inherent robustness against partial visibility and occlusions. This is particularly advantageous for systems containing event-based sensors where data streams may be spatially sparse or temporally incomplete. Furthermore, by leveraging a target already proven in the frame-based domain, the proposed method can utilize established, high-performance implementations for pattern detection and calibration. This ensures a stable geometric reference for both modalities even under suboptimal observation conditions.


Instead of relying on hardware-intensive approaches like LED-based ChArUco marker arrays~\cite{zhang_improved_2023}, our proposed method utilizes a standard display for pattern presentation, effectively eliminating the need for specialized hardware. By capturing the ChArUco pattern concurrently with both the event and frame-based sensors, we establish a shared observation that serves as the basis for a unified intrinsic and extrinsic calibration framework.

\paragraph{Overlay Generation.}
%

To achieve this shared observation under static conditions, we temporally modulate the calibration pattern on a screen by blending~\cite{porter_compositing_1984} it with a uniform white image, as shown in Fig.~\ref{fig:method_overview}~(a). Let $I_{p}(\mathbf{x})$ denote the static ChArUco pattern and $I_{w}(\mathbf{x})$ a constant white image of identical dimensions. We define a blended state $I_{\alpha}(\mathbf{x})$ as a linear interpolation between these two images:
\begin{equation}
\label{eq:blended_state}
I_{\alpha}(\mathbf{x}) = (1-\alpha)~I_{p}(\mathbf{x}) + \alpha~I_{w}(\mathbf{x}),
\end{equation}
where $\alpha \in (0,1)$ is a fixed opacity parameter. To induce events, the display stimulus $I(\mathbf{x}, t)$, where $t$ denotes the continuous time variable, periodically alternates between the static ChArUco pattern $I_{p}(\mathbf{x})$ and the blended state $I_{\alpha}(\mathbf{x})$, as illustrated in Fig.~\ref{fig:method_overview}~(b). We define a full modulation cycle at a frequency of $f = 15$~Hz, corresponding to $30$ state transitions per second, yielding the following temporal profile:
\begin{equation}
\label{eq:temporal_switching}
I(\mathbf{x}, t) = 
\begin{cases} 
I_{p}(\mathbf{x}) & \text{if } \lfloor 2 f t \rfloor \bmod 2 = 0 \\
I_{\alpha}(\mathbf{x}) & \text{if } \lfloor 2 f t \rfloor \bmod 2 = 1.
\end{cases}
\end{equation}
This discrete switching creates sufficient log-intensity changes $\Delta \ln I(\mathbf{x},t)$ at $33.3$~ms intervals to reliably trigger events. Crucially, selecting an appropriate $\alpha$ ensures the ChArUco geometry remains continuously detectable to the frame-based sensor. By avoiding the blank states characteristic of binary flickering, this persistent visibility eliminates the need for strict hardware-level synchronization between the display device and the cameras. Thus, the target can be pre-rendered as a standard video sequence and displayed on any commodity hardware, such as a smartphone, tablet, or monitor.

%

\paragraph{Data Processing.}
To establish a weak correspondence between the sensing modalities, both sensors are temporally aligned using a lightweight, software-based scheme that provides only coarse synchronization. The frame-based camera serves as the master device, providing trigger signals for the event stream. No precise or hardware-level synchronization is enforced, making the setup straightforward to implement and requiring no additional synchronization hardware. While this results in temporal offsets and potential jitter, the static nature of our proposed method renders the calibration insensitive to such inaccuracies.

For feature extraction, the asynchronous event stream is mapped into a discrete 2D representation. As shown in Fig.~\ref{fig:method_overview}~(c), we construct event frames by accumulating events over temporal intervals aligned with consecutive captures of the frame-based camera~\cite{gallego_event-based_2022}. Both the generated event frames and the corresponding RGB images are converted to grayscale for further processing.

To ensure reliable detection of ArUco markers within the ChArUco board, a median filter is applied to the event frames. This suppresses noise and densifies the sparse regions of the calibration pattern while preserving the underlying geometric structure. These processed frames are then integrated into a standard calibration pipeline (e.g., OpenCV~\cite{opencv_library}) to solve for the final intrinsic and stereo extrinsic parameters, as outlined in Fig.~\ref{fig:method_overview}~(c–e). The resulting pipeline is modality-agnostic and robust, supporting both handheld acquisition and fixed-base robot-assisted setups. In robot-assisted configurations, the system actuates either the camera rig or the calibration target through a sequence of perspectively diverse poses, triggering captures for both sensors only once a stationary scene is ensured.

%% file: sec/4_experiments.tex
\section{Experiments}
\label{sec:Experiments}
A series of experiments was conducted to analyze the operating range, limitations, and robustness of the proposed cross-modal stereo calibration approach. The ablation studies focus on parameter influences related to image overlaying, sensitivity to display properties, statistical stability, illumination stability, viewing angle stability, and the ease of operation in practical, real-world scenarios. To situate our work within the current state-of-the-art, we further compare against the motion-based E2Calib framework of Muglikar~\etal~\cite{muglikar_how_2021}, including its ChArUco-adapted and Kalibr-based variants, as well as the non-motion-based approach of Plasberg~\etal~\cite{plasberg_intrinsic_2022}. Finally, we demonstrate the practical utility of our framework through a downstream hand-eye calibration task on a robotic platform.

\subsection{Experimental Setup}
\label{sec:experimental_setup}
To ensure high repeatability and objective comparison across all evaluation scenarios, we utilize a standardized data collection pipeline.

\paragraph{Sensor Configuration.} 
Our sensor rig consists of a frame-based IDS UI-5250CP camera (native resolution $1600 \times 1200~\mathrm{px}$) and a Prophesee EVK4 event-based sensor (resolution $1280 \times 720~\mathrm{px}$). To ensure both sensors observe the same real‑world field of view, the UI-5250CP images were cropped to a region of interest of $1380 \times 780~\mathrm{px}$, matching the Prophesee sensor’s field of view, considering the smaller effective pixel size of the frame-based camera. Both sensors were equipped with identical $12\,\text{mm}$ C-mount lenses (IDS-2M23-C1214) that were manually focused prior to the experiments. The aperture was set to $f/11$ to provide a large depth of field, ensuring the calibration pattern remains sharp across a wide range of poses.

\paragraph{Calibration Settings.} 
The sensors were specifically tuned to the requirements of their respective modalities. For the frame-based camera, parameters like exposure time and gain were adjusted to achieve a well-exposed image, while the bias settings of the event sensor were optimized to produce a dense, but noise-reduced event stream. Once established, these settings were frozen for all subsequent experiments to maintain strict comparability. A detailed overview of the parameter configurations is provided in Appendix~\ref{sec:experimental_setup_appendix}. Parameters not explicitly listed in Tab.~\ref{tab:appendix_ueye_settings} and Tab.~\ref{tab:appendix_sensor_settings} were kept at their manufacturer hardware default settings.

\paragraph{Robot and Display Configuration.} 
All experiments, except the handheld evaluation, were performed using a Universal Robots UR5e 6-DOF manipulator. For the automated primary ablation studies the robot moved the sensor rig through a trajectory of $39$ diverse poses relative to a static $27$-inch Dell S2722QC monitor ($3840 \times 2160~\mathrm{px}$). At each pose, the robot remained stationary during the capture sequence to ensure a stable acquisition for both modalities and to prevent motion-induced artifacts. 

For the comparative analysis against the method of Muglikar~\etal, the same $39$ poses were utilized. However, in accordance with the procedure described in the respective original work, the data was acquired while the robot with the camera rig was in motion. An overview of the setup used for the robot-assisted ablation studies and comparative analysis is provided in Fig.~\ref{fig:experimental_setup} of Appendix~\ref{sec:experimental_setup_appendix}. 

In addition to the robot-based configuration, a handheld setup was evaluated using the OLED display of an iPhone~15 ($2556 \times 1179~\mathrm{px}$) to assess the applicability across different display technologies. In this case, the calibration target was moved manually across at least $39$ poses, resulting in less repeatable but more unconstrained viewpoint distributions compared to the robot-based setup. 

For the hand-eye calibration task, the configuration was adapted. A compact $3.5$-inch Waveshare LCD ($480 \times 800~\mathrm{px}$) was mounted to the robotic arm as a mobile calibration target. The robot traversed $22$ distinct poses or $6$ trajectories in both scenarios, respectively, within the static sensor rig's field of view.

\subsection{Ablation Studies}

\begin{figure}[t]
\centering
\includegraphics[width=1.0\linewidth]{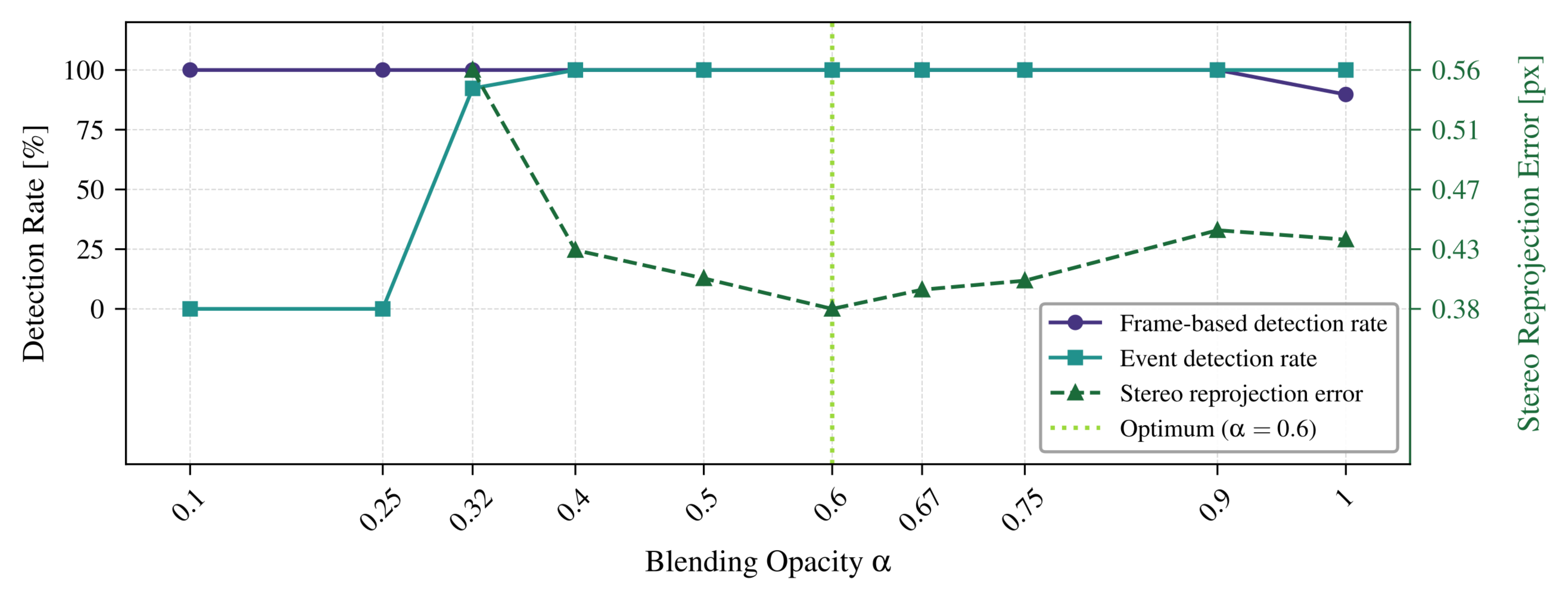}
\caption{Detection rate and stereo reprojection error as a function of the blending opacity $\alpha$. Frame-based detection remains close to $100\%$ across most opacity levels, with a slight degradation for $\alpha \rightarrow 1$ due to the sporadic invisibility of the pattern in the absence of temporal display synchronization. In contrast, event-based detection increases with $\alpha$ and saturates for $\alpha \geq 0.4$. The stereo reprojection error reaches a minimum at $\alpha = 0.6$, indicating an optimal trade-off between detection robustness and calibration accuracy.}
\label{fig:alpha_sweep}
\end{figure}
\paragraph{Opacity Parameter.} 
The influence of the blending opacity $\alpha$ on cross-modal calibration performance was investigated through a comprehensive ablation study to determine a value that ensures high fidelity for both sensing modalities.

The evaluation shown in Fig.~\ref{fig:alpha_sweep} reveals that the operating point must be carefully selected to satisfy the conflicting sensor requirements. At lower values ($\alpha < 0.32$), the intensity variations produced by the toggling overlay are insufficient to generate a reliable stream of events, resulting in a total failure of the event-based detection pipeline. Conversely, higher alpha values improve event triggering at the cost of increased occlusion of the calibration pattern during overlay for the frame-based sensor. This effect degrades the frame‑based camera, causing a measurable drop in detection rate and an increase in stereo reprojection error as $\alpha$ approaches $1.0$. Since there is no synchronization between the frame-based sensor and the display, extreme overlay settings can arbitrarily lead to frames in which the calibration pattern is not visible for the sensor.

The study confirms that the system achieves its most balanced performance at $\alpha = 0.6$. At this operating point, the induced intensity changes are sufficient to ensure a $100\%$ event detection rate, while maintaining adequate transparency for precise frame‑based calibration target detection, resulting in the minimum observed stereo reprojection error of $0.38$~px. Consequently, $\alpha = 0.6$ is used for all subsequent experiments unless stated otherwise.


\begin{table}[t]
\centering
\caption{Influence of display brightness on stereo calibration performance for a single experimental run, evaluated across brightness levels from $0\%$ to $100\%$. The table reports the measured luminance of the white ($L_{\text{white}}$) and black ($L_{\text{black}}$) display states, the combined successful corner detection rate ($u_c$), and the stereo reprojection error ($\varepsilon_{\text{rep}}$), with arrows ($\uparrow, \downarrow$) indicating the direction of improved performance.}
\vspace{4pt}
\begin{tabular}{|c|c|c|c|c|}
\hline
Brightness [\%] &
$L_{\text{white}}$ [cd/m$^2$] &
$L_{\text{black}}$ [cd/m$^2$] &
$u_c$ [\%] $\uparrow$ &
$\varepsilon_{\text{rep}}$ [px] $\downarrow$ \\
\hline\hline
0   & 35.0  & 0.03  & 15.3 & -- \\
20  & 78.0  & 0.06  & 98.5 & 0.442 \\
40  & 121.0 & 0.10  & 100.0 & 0.418 \\
60  & 164.0 & 0.13  & 100.0 & 0.405 \\
80  & 236.0 & 0.19  & 100.0 & 0.395 \\
100 & 391.0 & 0.32  & 100.0 & 0.384 \\
\hline
\end{tabular}
\label{tab:display_brightness_ablation}

\vspace{2pt}
{\footnotesize -- Indicates that insufficient corner detections precluded calibration.}
\end{table}

\paragraph{Display Brightness.} 
To evaluate the versatility of our framework across different display hardware, we varied the brightness of our test monitor from $0\%$ to $100\%$ (Tab.~\ref{tab:display_brightness_ablation}). While the contrast ratio remains stable at approximately $1250:1$, absolute luminance significantly impacts both sensing modalities. At $0\%$ brightness, the RGB sensor suffers from severe underexposure, while the event camera's signal-to-noise ratio (SNR) is heavily degraded. Combined, these factors result in a $15.3\%$ corner detection rate, preventing stereo calibration. Notably, for all brightness settings of $20\%$ and above, the system achieves a success rate exceeding $98\%$. The reprojection error further improves from $0.442$~px to $0.384$~px as brightness increases, likely due to increased pixel bandwidth and reduced temporal jitter in the event stream. These results demonstrate that our proposed method is robust across a wide range of displays. Unless indicated otherwise, a display brightness of $80\%$ was used in all subsequent experiments, providing low stereo reprojection errors and a default luminance level representative of state-of-the-art consumer displays.


\begin{table}[t]
\centering
\caption{Statistical robustness evaluation of the intrinsic parameters over 20 independent calibration runs using a robotic setup. The table reports the mean ($\mu$) and standard deviation ($\sigma$) for the corner detection success rate ($u_c$), focal lengths ($f_x, f_y$), principal point ($c_x, c_y$), and reprojection error ($\varepsilon_{\text{rep}}$).}
\vspace{4pt}
\begin{tabular}{|lr|c|c|c|c|c|c|}
\hline
Sensor & &
$u_c$ [\%] $\uparrow$ &
$f_x$ [px]&
$f_y$ [px]&
$c_x$ [px]&
$c_y$ [px]&
$\varepsilon_{\text{rep}}$ [px] $\downarrow$ \\
\hline\hline
RGB & ($\mu$) & 100.0 & 2771.5 & 2772.6 & 800.1 & 450.8 & 0.142 \\
& ($\sigma$) & 0.0 & 3.5 & 3.4 & 1.8 & 1.4 & 0.006 \\
\hline
Event & ($\mu$) & 100.0 & 2571.1 & 2572.3 & 609.3 & 378.2 & 0.449 \\
& ($\sigma$) & 0.0 & 26.4 & 26.3 & 10.7 & 9.1 & 0.013 \\
\hline
\end{tabular}
\label{tab:mono_ablation_robustness}
\end{table}


\begin{table}[t]
\centering
\caption{Statistical robustness evaluation of the extrinsic parameters over 20 runs. The table reports the successful corner detection rate ($u_c$), the magnitude of the estimated rotation ($\|\mathbf{r}\|$) and translation ($\|\mathbf{t}\|$), and the reprojection error ($\varepsilon_{\text{rep}}$).}
\vspace{4pt}
\begin{tabular}{|lr|c|c|c|c|}
\hline
Sensor & &
$u_c$ [\%] $\uparrow$ &
$\|\mathbf{r}\|$ [\textdegree] &
$\|\mathbf{t}\|$ [m] &
$\varepsilon_{\text{rep}}$ [px] $\downarrow$ \\
\hline\hline
RGB + Event & ($\mu$) & 99.8 & 1.077 & 0.0346 & 0.383 \\
& ($\sigma$) & 0.5 & 0.193 & 0.00017 & 0.012 \\
\hline
\end{tabular}
\label{tab:stereo_ablation_robustness}
\end{table}

\paragraph{Calibration Repeatability.}

To evaluate the numerical robustness of the proposed method, we conducted $20$ independent calibration runs under identical conditions. As summarized in Tab.~\ref{tab:mono_ablation_robustness}, both sensors achieved a $100.0\,\%$ corner detection success rate, confirming the reliability of the feature acquisition process. The RGB camera intrinsics exhibited minimal variability, whereas the event camera displayed higher standard deviations due to the inherent sparsity and noise of event-based data. Nevertheless, it still maintained stable reprojection errors across all trials. Stereo calibration results (Tab.~\ref{tab:stereo_ablation_robustness}) further validate the repeatability of the framework. The estimated translation magnitude consistently aligned with the nominal baseline of $0.0344\,\text{m}$, derived from the camera rig's CAD model, with negligible variance. Rotation estimates, quantified by the angular magnitude $\|\mathbf{r}\|$, were highly consistent across repetitions. Their absolute accuracy is subject to mechanical installation tolerances of the camera mounts, which preclude an exact ground-truth comparison. Nevertheless, the consistently low reprojection errors across all phases underscore the reliability of the proposed framework.


\begin{table}[t]
\centering
\caption{Illumination robustness evaluation of the extrinsic parameters over 5 runs with varying spotlight positions. The table reports the successful corner detection rate ($u_c$), the magnitude of the estimated rotation ($\|\mathbf{r}\|$) and translation ($\|\mathbf{t}\|$), and the reprojection error ($\varepsilon_{\text{rep}}$).}
\vspace{4pt}
\begin{tabular}{|lr|c|c|c|c|}
\hline
Sensor & &
$u_c$ [\%] $\uparrow$ &
$\|\mathbf{r}\|$ [\textdegree] &
$\|\mathbf{t}\|$ [m] &
$\varepsilon_{\text{rep}}$  [px] $\downarrow$ \\
\hline\hline
RGB + Event & ($\mu$) & 92.8 & 1.252 & 0.0345 & 0.402 \\
& ($\sigma$) & 10.8 & 0.266 & 0.00012 & 0.017 \\
\hline
\end{tabular}
\label{tab:stereo_ablation_illumination}
\end{table}

\paragraph{Illumination Stability.} 
The robustness of the framework to adverse lighting was evaluated through $5$ calibration trials under artificial spotlight interference. We varied the position of a high-intensity LED source across $5$ distinct locations, including lateral and vertical offsets, to simulate challenging real-world conditions such as direct sunlight. For reference, a qualitative example of the applied illumination disturbance is shown in Fig.~\ref{fig:illumination_samples} of Appendix~\ref{sec:appendix_intrinsic_calibration}. The resulting extrinsic statistics are detailed in Tab.~\ref{tab:stereo_ablation_illumination}. An analysis of the corresponding intrinsic calibration behavior under these illumination conditions is provided in Appendix~\ref{sec:appendix_intrinsic_calibration}, but is not the primary focus of this study.

Our extrinsic calibration results indicate that while strong, spatially varying illumination noticeably reduces the corner detection success rate and increases its variability, the geometric integrity of the calibration remains intact. Specifically, the estimated translation magnitude remains consistent with the nominal baseline, averaging $0.0345\,\text{m}$ across all disturbed trials. While the rotation estimates and reprojection errors show a marginal increase in variance compared to controlled conditions, they remain within a low-error regime. These findings demonstrate that the chosen pattern and proposed calibration method are resilient to localized illumination artifacts. Disturbances primarily affect detection efficiency rather than final registration accuracy. Such robustness confirms the potential suitability of the method for environments where controlled illumination cannot be guaranteed.

\begin{table}[t]
\centering
\setlength{\tabcolsep}{4pt} 
\caption{Viewing angle stability evaluation across viewing angles from $30^\circ$ to $70^\circ$, evaluated along the vertical (V), horizontal (H), and diagonal (D) axes. The table reports the combined corner detection rate ($u_c$) and reprojection error ($\varepsilon_{\text{rep}}$), averaged over 5 runs per direction.}
\vspace{4pt}
\label{tab:stereo_ablation_angle}
\begin{tabular}{|lr|c|c|c|c|c|c|}
\hline
 & &
\multicolumn{2}{c|}{Vertical (V)} &
\multicolumn{2}{c|}{Horizontal (H)} &
\multicolumn{2}{c|}{Diagonal (D)} \\
\hline
Angle & & 
$u_c$ & $\varepsilon_{\text{rep}}$ & 
$u_c$ & $\varepsilon_{\text{rep}}$ & 
$u_c$ & $\varepsilon_{\text{rep}}$ \\
 & & 
[\%] $\downarrow$ & [px] $\uparrow$ & 
[\%] $\downarrow$ & [px] $\uparrow$ & 
[\%] $\downarrow$ & [px] $\uparrow$ \\
\hline\hline
30$^\circ$ & & 100.0 & 0.322 & 100.0 & 0.320 & 100.0 & 0.313 \\
40$^\circ$ & & 100.0 & 0.363 & 100.0 & 0.342 & 100.0 & 0.388 \\
50$^\circ$ & & 100.0 & 0.374 & 100.0 & 0.359 & 89.2 & 0.482 \\
55$^\circ$ & & 100.0 & 0.436 & 56.7 & 0.374 & 48.3 & 0.926 \\
60$^\circ$ & & 100.0 & 0.477 & 14.1 & 0.377 & 0.0 & -- \\
65$^\circ$ & & 96.7 & 0.606 & 0.0 & -- & 0.0 & -- \\
70$^\circ$ & & 25.8 & 0.791 & 0.0 & -- & 0.0 & -- \\
\hline
\end{tabular}

\vspace{2pt}
{\footnotesize -- Indicates that insufficient corner detections precluded calibration.}
\end{table}

\paragraph{Viewing Angle Stability.} 
To evaluate robustness against viewing angle variations, we conducted tests using the Dell monitor described in Sec.~\ref{sec:experimental_setup}. Here, the viewing angle denotes the angular deviation of the camera's optical axis from the display's surface normal. The diagonal axis corresponds to a combined tilt applied simultaneously and equally along the vertical and horizontal directions. As detailed in Tab.~\ref{tab:stereo_ablation_angle}, angles from $30^\circ$ to $70^\circ$ were evaluated along the vertical (V), horizontal (H), and diagonal (D) axes, averaged over 5 runs per orientation. The proposed method remains stable up to $50^\circ$, maintaining a corner detection rate of $u_c \geq 89.2\,\%$ alongside low reprojection errors across all axes. This range well exceeds typical calibration requirements. Beyond $50^\circ$, performance degrades sharply. This is most pronounced along the diagonal axis, where detection fails entirely from $60^\circ$ onward. Despite this degradation at extreme angles, our approach remains reliable across the range of poses relevant for practical calibration.

\paragraph{Handheld Calibration.}

\begin{table}[t]
\centering
\caption{Handheld calibration evaluation of the extrinsic parameters over 5 runs. The table reports the successful corner detection rate ($u_c$), the magnitude of the estimated rotation ($\|\mathbf{r}\|$) and translation ($\|\mathbf{t}\|$), and the reprojection error ($\varepsilon_{\text{rep}}$).}
\vspace{4pt}
\begin{tabular}{|lr|c|c|c|c|}
\hline
Sensor & &
$u_c$ [\%] $\uparrow$ &
$\|\mathbf{r}\|$ [\textdegree] &
$\|\mathbf{t}\|$ [m] &
$\varepsilon_{\text{rep}}$ [px] $\downarrow$ \\
\hline\hline
RGB + Event & ($\mu$) & 100.0 & 0.9435 & 0.0343 & 0.361 \\
& ($\sigma$) & 0.0 & 0.285 & 0.00027 & 0.041 \\
\hline
\end{tabular}
\label{tab:stereo_ablation_handheld}
\end{table}
To further assess the practical applicability of the proposed approach, we perform an ablation study using a handheld acquisition setup. In contrast to the robot-based evaluation, this configuration removes precise and repeatable motion control, resulting in more arbitrary pose distributions. For each run, at least $39$ poses are captured with similar spatial coverage as in the robot-based experiments. A total of $5$ independent runs are conducted.

The results are summarized in Tab.~\ref{tab:stereo_ablation_handheld} and compared to the statistical robustness evaluation in Tab.~\ref{tab:stereo_ablation_robustness}. In this handheld setting, our proposed method achieves a perfect corner detection rate in both modalities, indicating reliable feature extraction even under less constrained acquisition. Furthermore, the mean reprojection error is slightly lower than in the robot-based setup. This improvement is likely due to the higher contrast and smaller pixel pitch of the OLED display employed in the handheld setup, compared to the Dell LCD monitor. However, reduced pose control inherent to manual acquisition increases the standard deviation, though transformation magnitudes remain comparable.


 Overall, these results demonstrate that the proposed calibration approach generalizes well to practical scenarios, where precise positioning hardware may not be available.

\subsection{Comparative Evaluation}


\begin{figure}[t]
\centering
\begin{tabular}{cc}
\fbox{\includegraphics[width=0.44\linewidth]{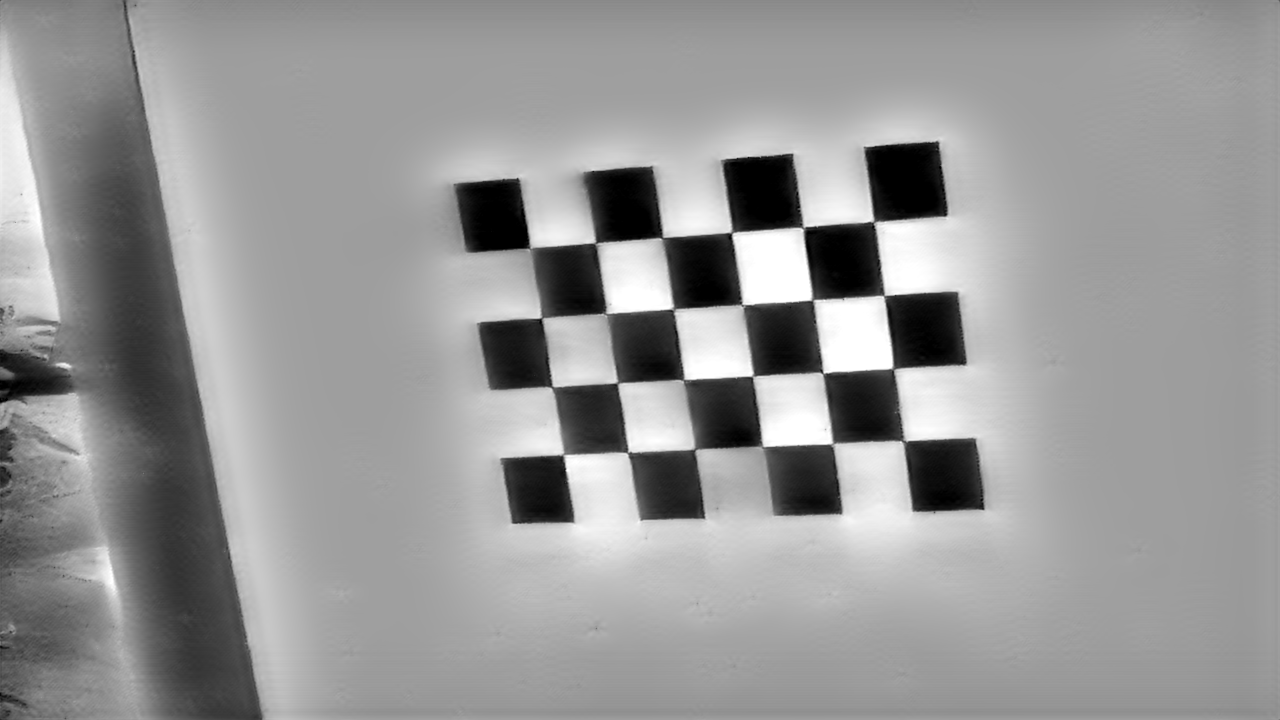}} &
\fbox{\includegraphics[width=0.44\linewidth]{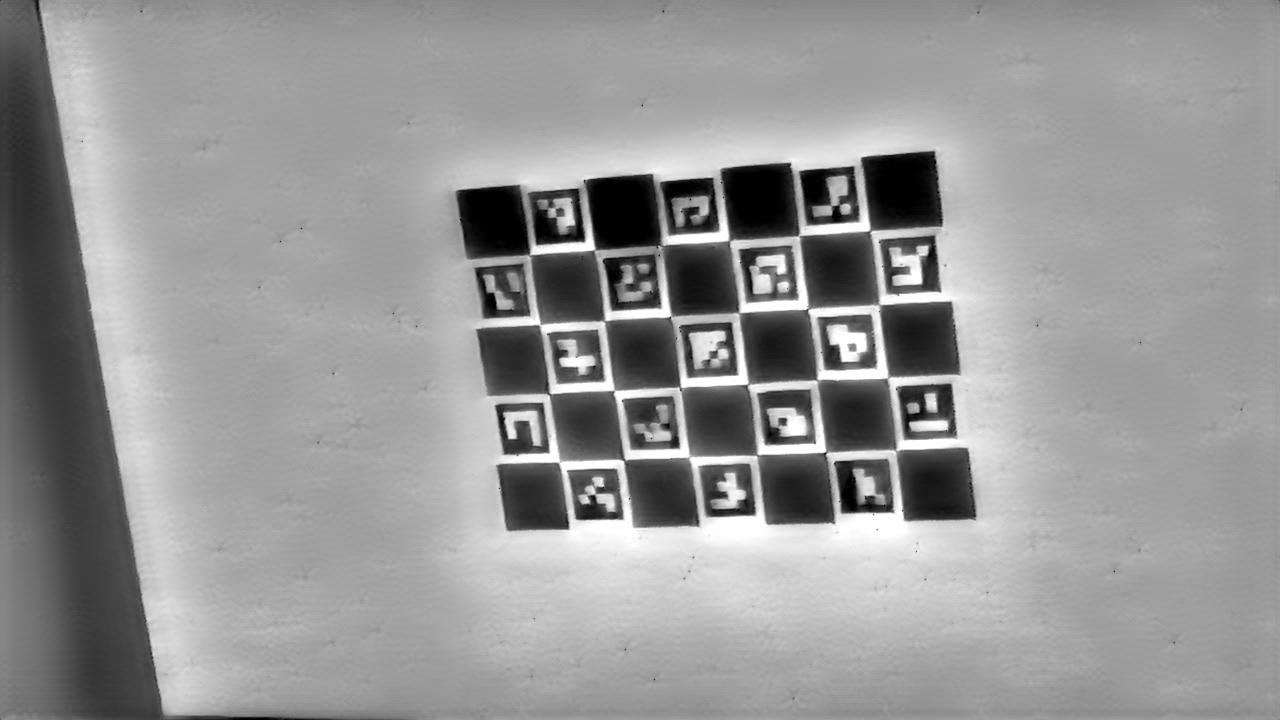}} \\
(a) Checkerboard E2VID Reconstruction  & (b) ChArUco E2VID Reconstruction 
\end{tabular}
\vspace{4pt}
\caption{E2VID reconstructed intensity frames used for the original E2Calib baseline (checkerboard) and our adapted ChArUco target-based variant.}
\label{fig:e2calib_methods}
\end{figure}


\begin{table}[t]
\centering
\caption{Extrinsic calibration performance of the motion-requiring baseline (E2Calib), its ChArUco-based adaptation (E2Calib + ChArUco), a variant using Kalibr instead of OpenCV for calibration (E2Calib + Kalibr), and the static baseline method of Plasberg et al., each evaluated over 5 runs. The table reports the successful corner detection rate ($u_c$), the magnitude of the estimated rotation ($\|\mathbf{r}\|$) and translation ($\|\mathbf{t}\|$), and the reprojection error ($\varepsilon_{\text{rep}}$).}
\vspace{4pt}
\begin{tabular}{|lr|c|c|c|c|}
\hline
Method & &
$u_c$ [\%] $\uparrow$ &
$\|\mathbf{r}\|$ [\textdegree] &
$\|\mathbf{t}\|$ [m] &
$\varepsilon_{\text{rep}}$ [px] $\downarrow$ \\
\hline\hline
E2Calib & ($\mu$) & 47.7 & 1.437 & 0.0339 & 1.235 \\
& ($\sigma$) & 5.6 & 0.386 & 0.0004 & 0.156 \\
\hline
E2Calib + ChArUco & ($\mu$) & 67.7 & 1.791 & 0.0338 & 1.161 \\
& ($\sigma$) & 3.7 & 0.381 & 0.0003 & 0.108 \\
\hline
E2Calib + Kalibr & ($\mu$) & * & 1.581 & 0.0360 & 0.687 \\
& ($\sigma$) & * & 0.178 & 0.0002 & 0.038 \\
\hline
Plasberg et al. & ($\mu$) & 56.4 & 2.039 & 0.0352 & 0.406 \\
& ($\sigma$) & 10.1  & 0.175 & 0.0001 & 0.019 \\
\hline
\end{tabular}
\label{tab:comparative_evaluations}

\vspace{2pt}
{\footnotesize * Values are not reported in the Kalibr calibration output.}
\end{table}

To assess the effectiveness of the proposed approach, we compare against four baselines. The first is the reconstruction-based E2Calib method of Muglikar \etal~\cite{muglikar_how_2021}. We additionally consider an adapted variant that employs a ChArUco calibration target instead of the checkerboard target (E2Calib + ChArUco, see Fig.~\ref{fig:e2calib_methods}), as well as a further variant substituting Kalibr for OpenCV as the calibration backend (E2Calib + Kalibr). Finally, we include the static approach of Plasberg \etal~\cite{plasberg_intrinsic_2022}. All baselines are evaluated over $5$ repetitions and reported in Tab.~\ref{tab:comparative_evaluations}, while the performance of our overlay method is summarized in Tab.~\ref{tab:stereo_ablation_robustness}.

Across all four baselines, our method consistently yields lower reprojection errors, indicating improved calibration accuracy. Among the E2Calib variants, the ChArUco adaptation shows a minor improvement in standard deviations and corner detection rates, both attributable to the ChArUco target, which allows frames with only partially detected corners to be used for calibration. Replacing OpenCV with Kalibr as backend, while retaining the checkerboard target, yields the lowest reprojection error among motion-based references, likely due to Kalibr's advanced optimization routines. Despite investigating these variations, our method reduces the mean reprojection error by $44\%$ relative to the best motion-based approach. This gap can be attributed to motion blur in the frame-based observations, together with the absence of strict hardware-level temporal synchronization. Compared to the static baseline of Plasberg \etal, our method still achieves a $6\%$ reduction in reprojection error.

As the results demonstrate, relying on a temporally modulated blended target substantially reduces hardware dependencies and deployment effort without sacrificing calibration accuracy. Our approach achieves superior accuracy compared to both motion-based and non-motion-based approaches.

\subsection{Case Study: Eye-to-Hand Calibration}
\begin{figure}[t]
\centering
\includegraphics[width=0.5\linewidth]{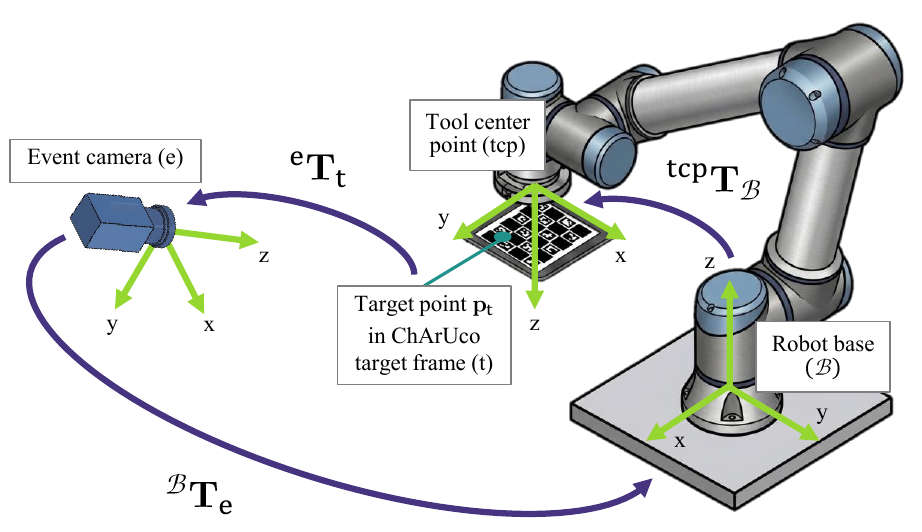}
\caption{Geometric transformation pipeline for eye-to-hand calibration, illustrating the mapping of a target point $\mathbf{p}_\mathrm{t}$ from the ChArUco target frame ($\mathrm{t}$) to the event camera frame ($\mathrm{e}$) via ${}^{\mathrm{e}}\mathbf{T}_\mathrm{t}$, and subsequently to the robot base frame ($\mathcal{B}$) using ${}^{\mathcal{B}}\mathbf{T}_\mathrm{e}$, with the final relation to the TCP defined through the robot kinematics ${}^{\mathrm{tcp}}\mathbf{T}_{\mathcal{B}}$.}
\label{fig:hec}
\end{figure}


\begin{table}[t]
\centering
\caption{Hand-eye calibration results aggregated over 20 valid runs. The table reports the mean ($\mu$) and standard deviation ($\sigma$) for the estimated translation vector ($\mathbf{t} = (t_x, t_y, t_z)$) between the event camera and robot base, its Euclidean norm ($\|\mathbf{t}\|$), and the rotation magnitude ($\theta$) obtained from the rotation matrix $\mathbf{R}$ via axis-angle representation.}
\vspace{4pt}
\begin{tabular}{|lr|c|c|c|c|c|}
\hline
Statistic & & $t_x$ [m] & $t_y$ [m] & $t_z$ [m] & $\|\mathbf{t}\|$ [m] & $\theta$ [\textdegree] \\ 
\hline\hline
${}^{\mathcal{B}}\mathbf{T}_\mathrm{e}$ & ($\mu$) & 0.1258 & -0.5678 & 0.3839 & 0.6970 & 173.76 \\
& ($\sigma$) & 0.0093 & 0.0076 & 0.0061 & 0.0056 & 2.80 \\
\hline
\end{tabular}
\label{tab:hec_results}
\end{table}


\begin{table}[t]
\centering
\caption{Spatial consistency evaluation results averaged over 5 runs, with each run consisting of 22 diverse poses for the target-to-TCP evaluation and 6 trajectories for the base-frame distance evaluation. The table reports the mean Euclidean distance ($\overline{d}$) between the transformed calibration point (ID9) and the TCP, its standard deviation ($\sigma_d$), the ground truth distance ($d_{\mathrm{gt}}$), and the mean absolute error ($\mathrm{MAE}$) with respect to the ground truth.}
\vspace{4pt}
\begin{tabular}{|l|c|c|c|c|}
\hline
Evaluation Mode & $\overline{d}$ [mm] & $\sigma_d$ [mm] $\downarrow$ & $d_{\mathrm{gt}}$ [mm] & $\mathrm{MAE}$ [mm] $\downarrow$ \\
\hline\hline
Target-to-TCP& 67.09 & 1.86 & 67.80 & 0.71 \\
Base-Frame Distance & Scene dep. & 0.66 & Scene dep. & 1.40 \\
\hline
\end{tabular}
\label{tab:hec_spatial_consistency}
\end{table}

To demonstrate that the proposed approach extends beyond cross-modal stereo calibration, we evaluate its applicability to downstream event-camera-based tasks. As a representative use case, we consider eye-to-hand calibration, depicted in Fig.~\ref{fig:hec}. We show that combining the ChArUco-based target with temporally modulated pattern blending enables reliable geometric estimation even under partial occlusions by the robot actuator (Appendix~\ref{sec:appendix_hec}).

To this end, we perform a spatial consistency analysis by selecting the event camera frame ($\mathrm{e}$) as the primary reference. The transformation ${}^{\mathcal{B}}\mathbf{T}_{\mathrm{e}} \in SE(3)$, mapping points from the event camera frame to the robot base frame ($\mathcal{B}$), is obtained via hand-eye calibration~\cite{tsai_new_1989} in an eye-to-hand configuration with a static camera and a movable calibration target. The ground-truth transformation was obtained from the CAD model of the experimental setup and additionally verified through manual measurements.

The estimated translation components, shown in Tab.~\ref{tab:hec_results}, are in close agreement with the ground truth ($t_x = 0.125\,\mathrm{m}$, $t_y = -0.563\,\mathrm{m}$, $t_z = 0.382\,\mathrm{m}$). Furthermore, the low standard deviation across all components indicates a stable estimation of the transformation parameters. Overall, the obtained results provide robust geometric alignment suitable for standard vision-guided robotic manipulation tasks. The estimated rotation is also consistent with the experimental setup, as the camera is approximately facing downward toward the robot base.


\paragraph{Target-to-TCP Consistency Evaluation.}
To assess the spatial consistency of the estimated transformations, we evaluate the stability of a reference point attached to the calibration target across multiple robot poses. Specifically, we consider a ChArUco corner with fixed identifier (ID9), which is reliably detectable (unobstructed) in all observations.


For each pose, the point is first reconstructed in the event camera frame as $\mathbf{p}_\mathrm{e} \in \mathbb{R}^3$ using pose estimation~\cite{lepetit_epnp_2009}. It is then transformed into the robot base frame using the previously estimated transformation:
\begin{equation}
\mathbf{p}_{\mathcal{B}} = {}^{\mathcal{B}}\mathbf{T}_\mathrm{e} \, \mathbf{p}_\mathrm{e},
\end{equation}
where ${}^{\mathcal{B}}\mathbf{T}_\mathrm{e} \in SE(3)$ denotes the previously established transformation. Subsequently, the point is expressed in the tool center point (TCP) frame using the known forward kinematics:
\begin{equation}
\mathbf{p}_{\mathrm{tcp}} = {}^{\mathrm{tcp}}\mathbf{T}_{\mathcal{B}} \, \mathbf{p}_{\mathcal{B}},
\end{equation}
where ${}^{\mathrm{tcp}}\mathbf{T}_{\mathcal{B}} = ({}^{\mathcal{B}}\mathbf{T}_{\mathrm{tcp}})^{-1}$ is obtained from the robot model. Finally, the Euclidean deviation between the transformed point and the TCP origin is computed:
\begin{equation}
d = \left\| \mathbf{p}_{\mathrm{tcp}} \right\|_2.
\end{equation}
For a consistent calibration, the distance $d$ should remain approximately constant across robot poses, since the relative geometry between the calibration target and gripper is fixed. Deviations from constancy indicate inaccuracies in the overall transformation chain, including errors in the hand-eye calibration, intrinsic camera parameters, and pose estimation.

The first row in Tab.~\ref{tab:hec_spatial_consistency} summarizes the results of the proposed spatial consistency evaluation. The estimated mean distance $\overline{d}$ closely matches the ground truth distance $d_{\mathrm{gt}}$, resulting in a low mean absolute error of $0.71\,\mathrm{mm}$, corresponding to a relative error of approximately $1.05\%$. Furthermore, the small standard deviation $\sigma_d$ indicates high consistency across different robot poses. These results suggest that the proposed calibration approach yields both accurate and stable geometric alignment, with minimal pose-dependent variation.

\paragraph{Base-Frame Distance Consistency Evaluation.}
As a complementary validation, we evaluate distance preservation in the robot base frame using the event camera as the reference. The target point (ID9) is first reconstructed in the event camera frame as $\mathbf{p}_\mathrm{e} \in \mathbb{R}^3$ using pose estimation~\cite{lepetit_epnp_2009} and transformed into the robot base frame via the transformation ${}^{\mathcal{B}}\mathbf{T}_\mathrm{e}$ obtained via eye-to-hand calibration.

For two target points $\mathbf{p}_{\mathrm{e},1}, \mathbf{p}_{\mathrm{e},2} \in \mathbb{R}^3$ with known distance $d_{\mathrm{gt}}$, the corresponding base frame distance is computed as
\begin{equation}
d_{\mathcal{B}} = \left\| {}^{\mathcal{B}}\mathbf{T}_\mathrm{e} \, \mathbf{p}_{\mathrm{e},1} - {}^{\mathcal{B}}\mathbf{T}_\mathrm{e} \, \mathbf{p}_{\mathrm{e},2} \right\|_2.
\end{equation}
The deviation from the known distance is defined as
\begin{equation}
e = \left| d_{\mathcal{B}} - d_{\mathrm{gt}} \right|.
\end{equation}
This evaluation is performed over $6$ scenarios. For a geometrically consistent calibration, $d_{\mathcal{B}}$ is expected to closely match $d_{\mathrm{gt}}$ across all observations. The results in the second row of Tab.~\ref{tab:hec_spatial_consistency} indicate that the base-frame distance consistency exhibits a low standard deviation $\sigma_d$, suggesting stable distance preservation across different trajectories. The low $\mathrm{MAE}$ further implies that the reconstructed distances deviate only slightly from the expected values despite being scene-dependent. These findings demonstrate that the proposed calibration approach reliably estimates the intrinsic parameters and provides a robust foundation for downstream tasks such as eye-to-hand calibration, resulting in stable and consistent transformation pipelines.

%% file: sec/5_limitations.tex
\section{Limitations}
\label{sec:Limits}
\paragraph{Display Properties.}
Despite the robustness of the proposed framework, its performance remains sensitive to the temporal characteristics of the display hardware. Specifically, event density scales with the refresh rate, potentially degrading marker detection accuracy below $30$~Hz. However, given that standard modern display hardware operates well above this range, this dependency poses a minimal constraint for real-world deployment. Furthermore, our proposed method assumes the use of a display with viewing-angle-stable contrast characteristics, as displays with strong angular dependence can exhibit reduced effective contrast modulation at large viewing angles. This particularly affects event cameras, which rely on intensity changes to register events.

\paragraph{Event Representation.} An additional limitation arises from the conversion of asynchronous event streams into discretized image representations. In the employed accumulation of events over time intervals, noise and temporal inconsistencies around high-contrast structures may introduce jitter, reducing corner localization accuracy and increasing reprojection error. More advanced reconstruction strategies, e.g., E2VID~\cite{rebecq_high_2021} or optimized accumulation schemes as in Wang \etal~\cite{wang_motion-error-free_2024}, could potentially mitigate these effects.


%% file: sec/6_conclusion.tex
\section{Conclusion}
\label{sec:concl}
In this paper, we presented a practical stereo calibration approach for hybrid event and frame-based systems using a temporally modulated, blended ChArUco target rendered frame-sequentially on standard commercial displays. By triggering events while ensuring the pattern remains continuously detectable in the frame domain, this method enables static acquisition and minimizes reliance on complex motion, reconstruction, or precise hardware synchronization. Our experiments demonstrate robust feature detection, as well as stable intrinsic and extrinsic estimation across diverse display settings, viewing angles, varying illumination, and in both robot-assisted and handheld capture. Furthermore, our approach successfully transfers to downstream eye-to-hand calibration, supporting consistent geometric estimation under practical conditions. Future work will explore improved event accumulation techniques to mitigate limitations related to display temporal characteristics and event discretization.

%% file: sec/appendix.tex
\section{Experimental Setup}
\label{sec:experimental_setup_appendix}

\begin{figure}[ht!]
\begin{center}
\fbox{\includegraphics[width=0.5\linewidth]{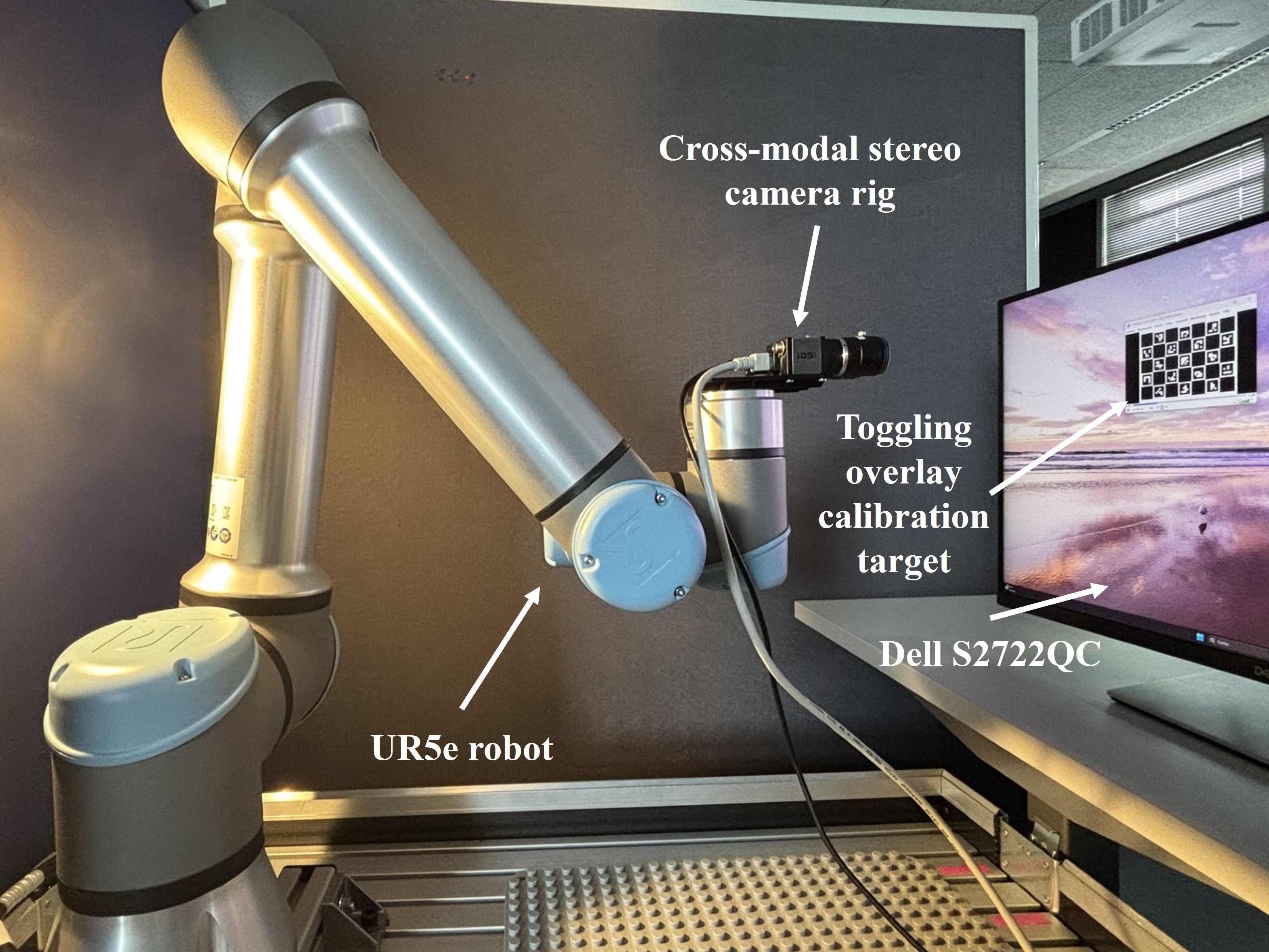}}
\end{center}
   \caption{The hybrid frame-event stereo rig is mounted on a UR5e robotic arm to ensure precise trajectory repeatability. The Dell S2722QC monitor displays the calibration pattern, serving as the cross-modal target for both the frame-based and event sensors.}
\label{fig:experimental_setup}
\end{figure}

\begin{table}[ht!]
\centering
\caption{IDS uEye frame-based camera configuration used in all experiments. The gamma value is specified in the camera parameterization scheme, where a value of 100 corresponds to a linear gamma of $\gamma=1.0$. All parameters not explicitly listed were kept at their hardware default settings.}
\vspace{4pt}
\begin{tabular}{|l|c|}
\hline
Parameter UI-5250CP & Value \\
\hline\hline
Exposure Time & $100.0\,\mathrm{ms}$ \\
Gain & 100  \\
Frame Rate & $10.0\,\mathrm{fps}$ \\
Gamma & 100 \\
\hline
\end{tabular}
\label{tab:appendix_ueye_settings}
\end{table}

\begin{table}[ht!]
\centering
\caption{Prophesee EVK4 sensor configuration used in all experiments. Bias values denote relative offsets from the hardware defaults and were tuned to optimize the signal-to-noise ratio for flickering pattern detection. All parameters not explicitly listed were kept at their hardware default settings.}
\vspace{4pt}
\begin{tabular}{|l|c|}
\hline
Parameter EVK4 & Value \\
\hline\hline
Event Trail Filter & Enabled \\
Filtering Type & \texttt{STC\_CUT\_TRAIL} \\
Trail Threshold & 100000 \\
\hline
\texttt{bias\_diff\_off} & $-15$ \\
\texttt{bias\_diff\_on} & $-35$ \\
\texttt{bias\_fo} & $-35$ \\
\texttt{bias\_hpf} & $80$ \\
\hline
\end{tabular}
\label{tab:appendix_sensor_settings}
\end{table}

\section{Ablation Illumination Stability}
\label{sec:appendix_intrinsic_calibration}

\begin{figure}[ht!]
\centering
\begin{tabular}{cc}
\fbox{\includegraphics[width=0.44\linewidth]{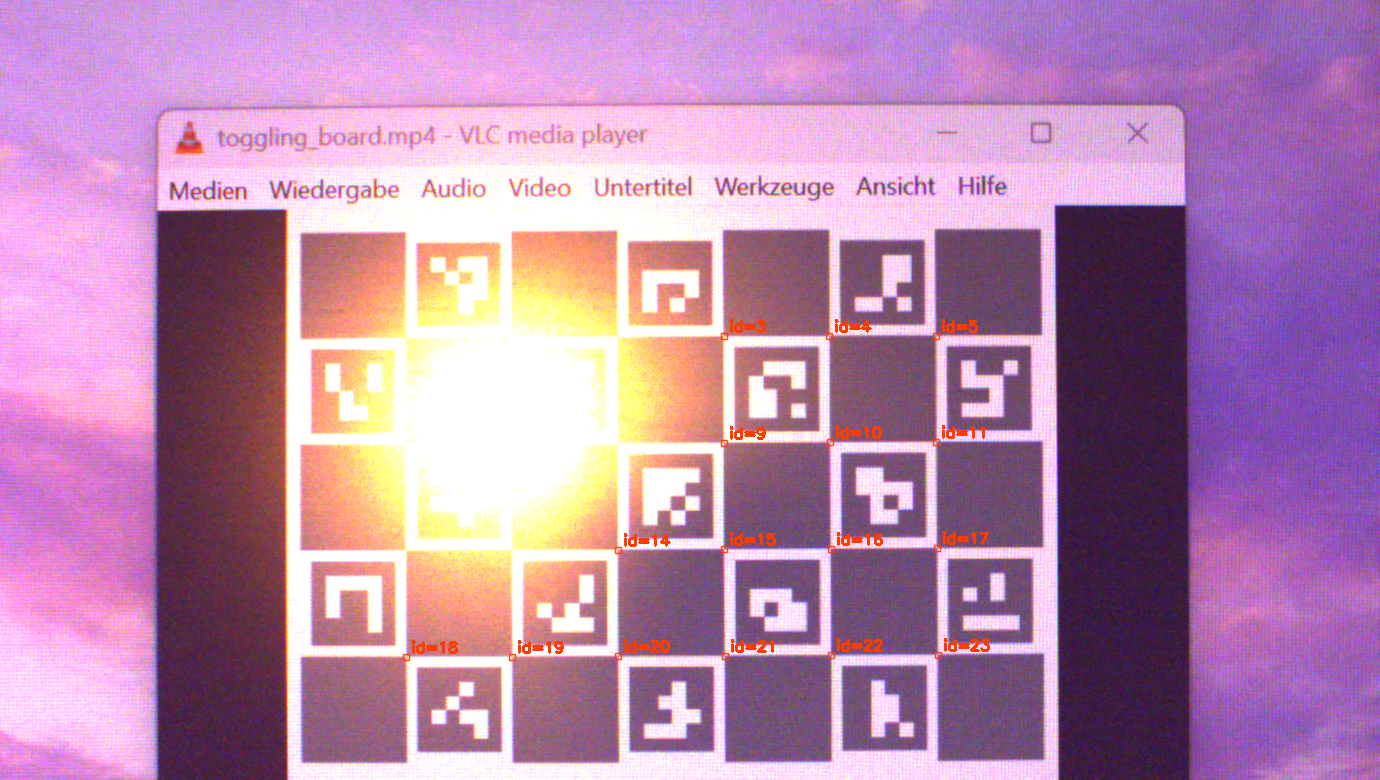}} &
\fbox{\includegraphics[width=0.44\linewidth]{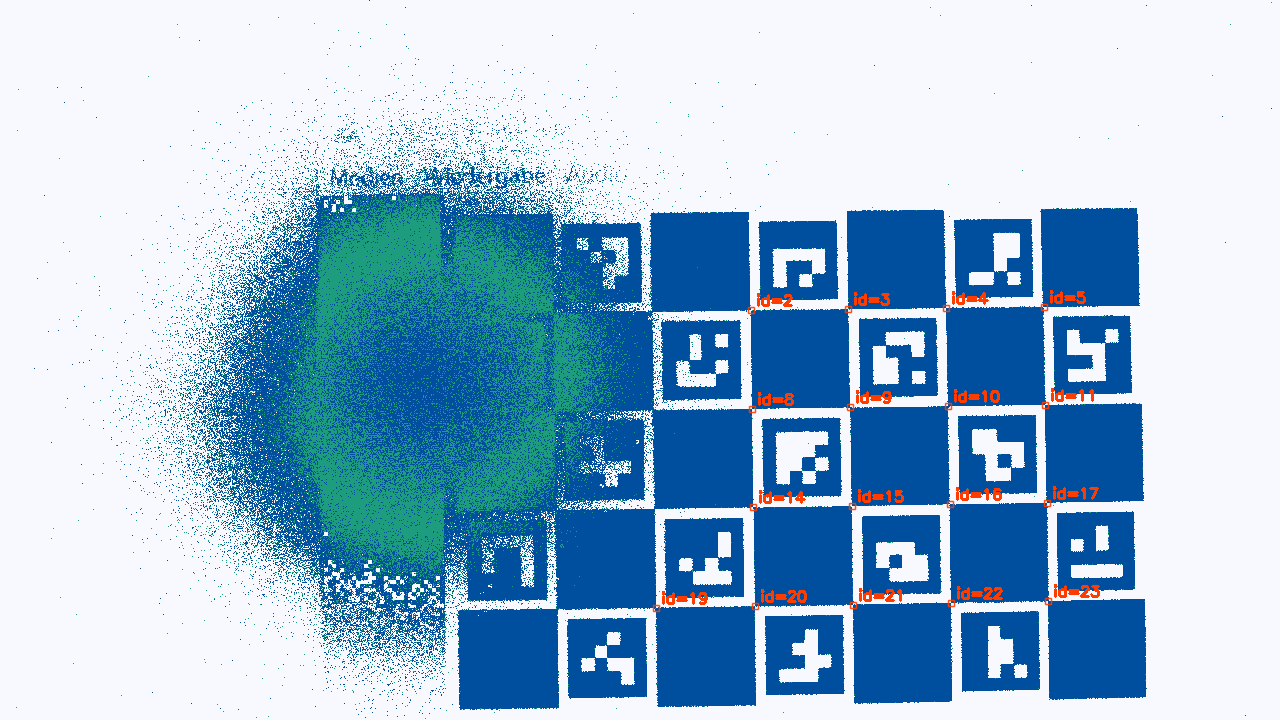}} \\
(a) RGB frame & (b) Event frame
\end{tabular}
\vspace{4pt}
\caption{Qualitative example from the illumination robustness experiment. Both images depict the same scene. Differences in perspective are due to the stereo baseline offset between the frame-based and event camera sensors.
(a) RGB image frame captured under strong external spotlight interference, causing overexposure on the calibration target.
(b) Corresponding event frame recorded under the same conditions, illustrating the effect of intense illumination on event generation and pattern visibility.}
\label{fig:illumination_samples}
\end{figure}


\begin{table}[ht!]
\centering
\caption{Illumination robustness evaluation of the intrinsic parameters with our approach over 5 runs with varying spotlight positions. The table reports the corner detection success rate ($u_c$), focal lengths ($f_x, f_y$), principal
point ($c_x, c_y$), and reprojection error ($\varepsilon_{\text{rep}}$).}
\vspace{4pt}
\begin{tabular}{|lr|c|c|c|c|c|c|}
\hline
Sensor & &
$u_c$ [\%] $\uparrow$ &
$f_x$ [px]&
$f_y$ [px]&
$c_x$ [px]&
$c_y$ [px]&
$\varepsilon_{\text{rep}}$ [px] $\downarrow$ \\
\hline\hline
RGB & ($\mu$) & 100.0 & 2764.8 & 2766.1 & 797.1 & 449.5 & 0.140 \\
& ($\sigma$) & 0.0 & 2.4 & 2.3 & 5.0 & 1.8 & 0.003 \\
\hline
Event & ($\mu$) & 98.5 & 2568.9 & 2569.6 & 600.5 & 367.8 & 0.456 \\
& ($\sigma$) & 3.4 & 11.4 & 11.6 & 9.3 & 10.2 & 0.007 \\
\hline
\end{tabular}
\label{tab:mono_ablation_illumination}
\end{table}

\section{Eye-To-Hand Calibration Occlusions}
\label{sec:appendix_hec}

\begin{figure}[ht!]
\centering
\begin{tabular}{cc}
\fbox{\includegraphics[width=0.44\linewidth]{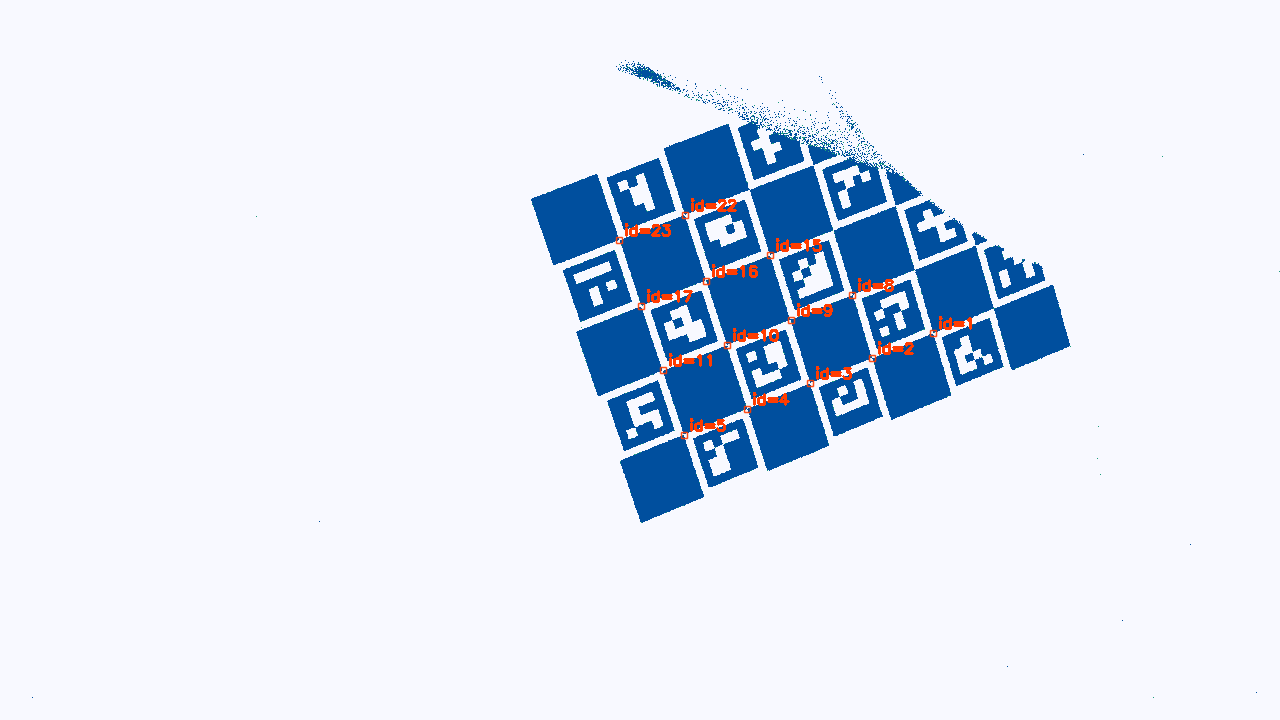}} &
\fbox{\includegraphics[width=0.44\linewidth]{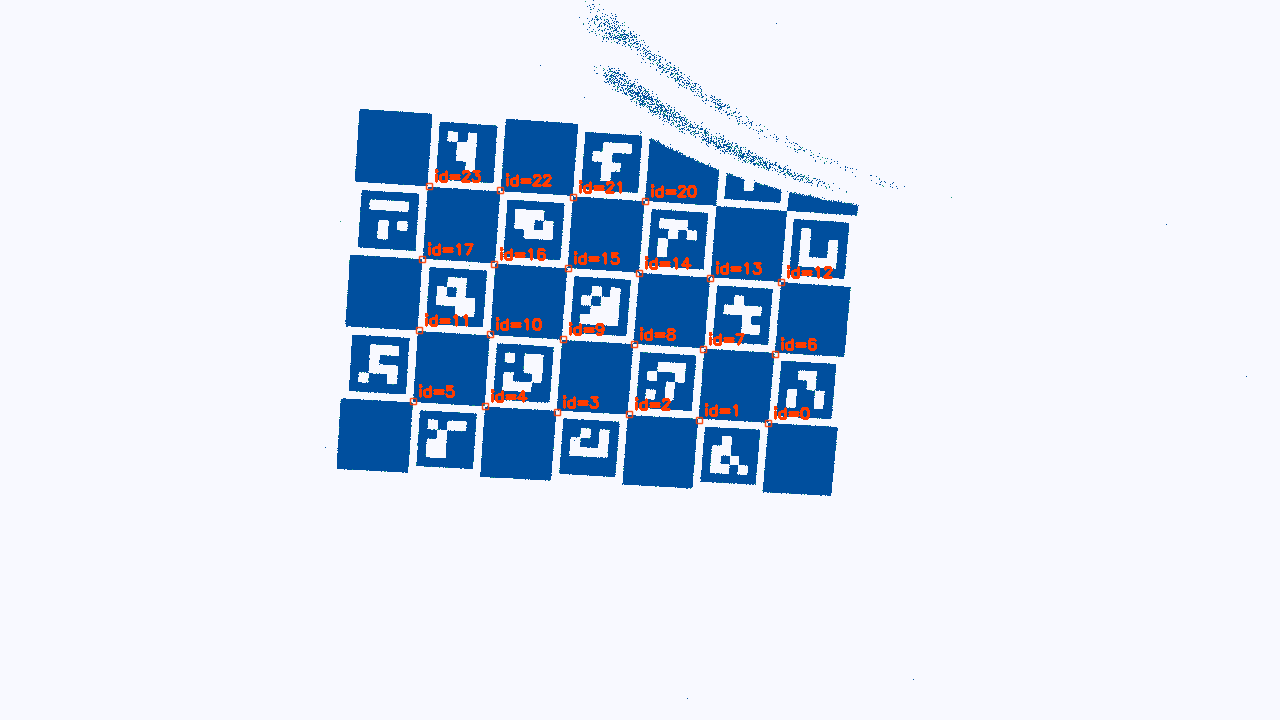}} \\
(a) Strong Occlusion & (b) Light Occlusion
\end{tabular}
\vspace{4pt}
\caption{Robustness of the proposed method under occlusion during eye-to-hand calibration. Even in the presence of partial occlusions caused by the robot actuator, salient ChArUco corners (highlighted in red) are reliably detected, enabling accurate calibration.}
\label{fig:hec_occlusion}
\end{figure}